\documentclass[10pt,twocolumn,letterpaper]{article}

\usepackage{iccv}
\usepackage{times}
\usepackage{graphicx}
\usepackage{amsmath}
\usepackage{amssymb}
\usepackage{booktabs}       
\usepackage{microtype}      
\usepackage[american]{babel}
\usepackage{graphicx}
\graphicspath{{figure/}}

\usepackage{multirow}
\usepackage{subcaption}

\def\pt{\phantom{0}}

\newcommand{\squishlist}{
 \begin{list}{$\bullet$}
  { \setlength{\itemsep}{0pt}
     \setlength{\parsep}{1pt}
     \setlength{\topsep}{1pt}
     \setlength{\partopsep}{0pt}
     \setlength{\leftmargin}{1em}
     \setlength{\labelwidth}{1em}
     \setlength{\labelsep}{0.5em} } }

\newcommand{\squishend}{
  \end{list}  }

\usepackage[pagebackref=true,breaklinks=true,colorlinks,bookmarks=false]{hyperref}

\iccvfinalcopy 

\ificcvfinal\fi

\begin{document}

\title{Rethinking Self-Supervised Learning: Small is Beautiful}

\author{Yun-Hao Cao and Jianxin Wu \\
National Key Laboratory for Novel Software Technology \\
Nanjing University, Nanjing, China \\
{\texttt{\{caoyunhao1997, wujx2001\}@gmail.com}}
}

\maketitle
\ificcvfinal\thispagestyle{empty}\fi

\begin{abstract}
Self-supervised learning (SSL), in particular contrastive learning, has made great progress in recent years. However, a common theme in these methods is that they inherit the learning paradigm from the supervised deep learning scenario. Current SSL methods are often pretrained for many epochs on large-scale datasets using high resolution images, which brings heavy computational cost and lacks flexibility. In this paper, we demonstrate that the learning paradigm for SSL should be different from supervised learning and the information encoded by the contrastive loss is expected to be much less than that encoded in the labels in supervised learning via the cross entropy loss. Hence, we propose scaled-down self-supervised learning (S3L), which include 3 parts: small resolution, small architecture and small data. On a diverse set of datasets, SSL methods and backbone architectures, S3L achieves higher accuracy consistently with much less training cost when compared to previous SSL learning paradigm. Furthermore, we show that even without a large pretraining dataset, S3L can achieve impressive results on small data alone. Our code has been made publically available at \url{https://github.com/CupidJay/Scaled-down-self-supervised-learning}.
\end{abstract}

\section{Introduction}

Deep supervised learning has achieved great success in the last decade. However, its dependency on image labels has driven people to explore a better solution. Self-supervised learning (SSL) has gained popularity because of its ability to avoid the cost of annotating large-scale datasets. After the emerging of the InfoNCE loss~\cite{InfoNCE:arxiv2018} and the contrastive learning paradigm, SSL has clearly gained momentum and a large amount of research contributions have been published, such as MoCo~\cite{moco:kaiming:CVPR20} (and MoCov2~\cite{mocov2:xinlei:arxiv2020}), SimCLR~\cite{simclr:hinton:ICML20} (and SimCLRv2~\cite{simclrv2:arxiv2020}), BYOL~\cite{byol:grill:NIPS20} and many more.

\begin{figure}
	\centering
	\includegraphics[width=\columnwidth]{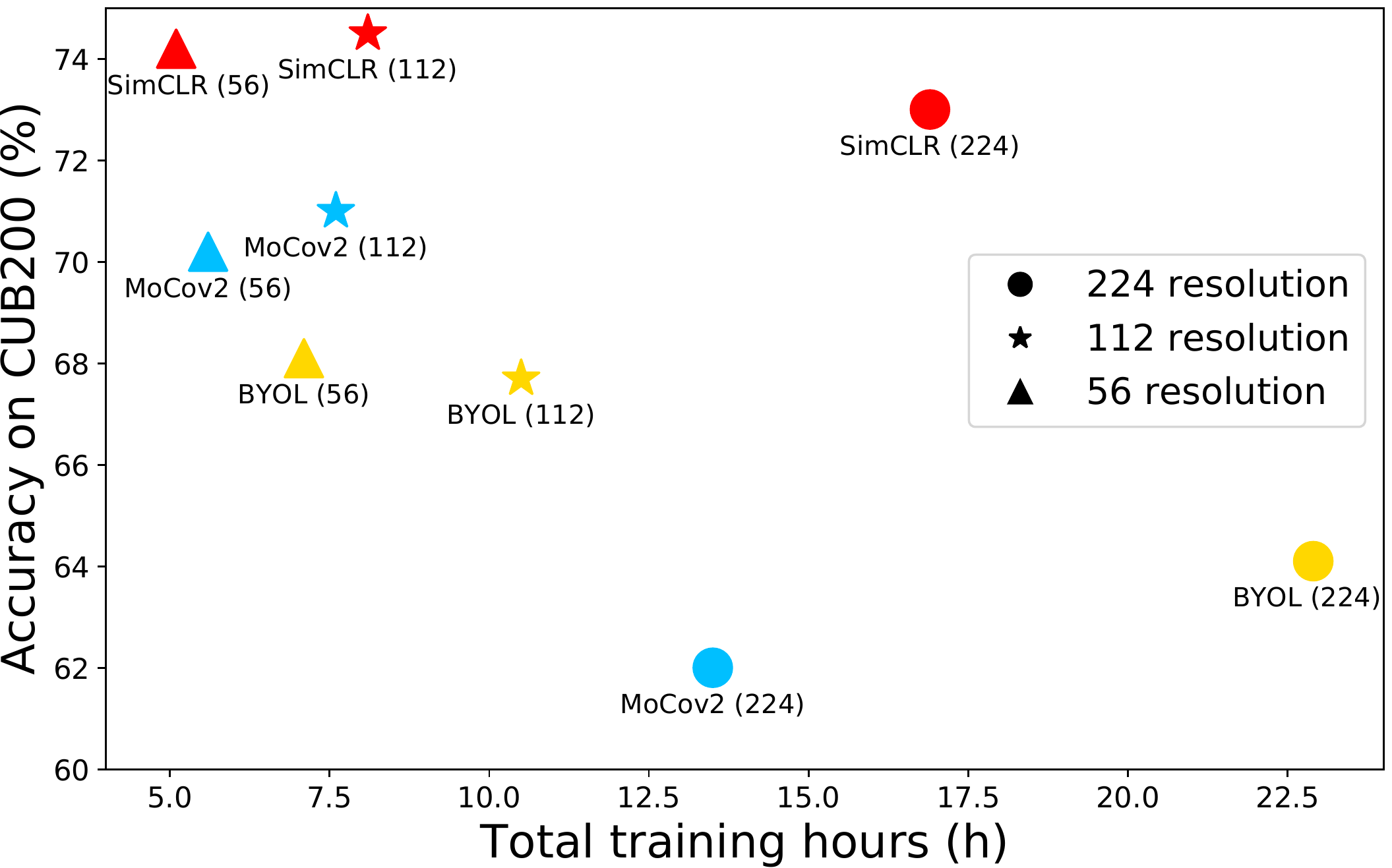}
	\caption{Accuracy of SSL methods on CUB200~\cite{cub200} with ResNet-50 backbone. \emph{Only} the small CUB200 dataset was used in SSL pretraining and subsequent fine-tuning. In each method (MoCov2, SimCLR and BYOL), smaller resolution not only trained much faster, but also achieved higher accuracy than the baseline 224 resolution.}
	\label{fig:scatter}
\end{figure}

A common theme in all these methods, however, is that \emph{they all learn self-supervised models in a setup that is clearly inherited from the supervised learning setting}. Common characteristics of these methods include: (1) Use 224x224 as the input resolution; (2) Use large-scale training sets (e.g., ILSVRC-2012~\cite{ILSVRC2012:russakovsky:IJCV15}); (3) Use the entire network architecture from supervised learning tasks (mostly ResNet-50~\cite{resnet:he:CVPR16}) and the entire backbone network afterwards in downstream tasks; (4) Require many training (e.g., 800 or more) epochs. The combination of these four characteristics dominates current SSL researches. This combination, however, has exhibited clear disadvantages in three important aspects:
\squishlist
	\item \textbf{Computing.} The combination of a large-scale dataset, a large input resolution, a large number of epochs and a complex backbone network means that SSL methods are computationally extremely expensive. This phenomenon makes SSL a privilege for researchers at few institutions.
	\item \textbf{Data.} SSL methods are often pretrained on a large-scale dataset (such as ILSVRC-2012 or even larger ones), and then fine-tuned in various downstream tasks. For a task where the \emph{total} amount of available images (labeled or not) is limited (e.g., 100 categories with roughly 20 images per category), it is unknown whether SSL will still be useful without a large pretraining dataset.
	\item \textbf{Flexibility.} This SSL paradigm (pretraining on big data followed by downstream fine-tuning) will sometimes become cumbersome. For instance, we need to train 10 different models for the same task, and deploy them to different hardware platforms~\cite{once-for-all:hansong:ICLR20}, but it is impractical to pretrain 10 models on a large-scale dataset.
\squishend

Self-supervised learning, however, dramatically differs from supervised learning. In SSL, why must we blindly inherit the setup from supervised learning? The most important difference is, of course, the presence or missing of image labels. As a direct consequence, we assume that \textit{the information encoded by the contrastive loss is expected to be much less than that encoded in the labels via the cross entropy loss}, which is the fundamental assumption of this paper. To properly learn lesser information, we rethink how SSL should be carried out, and recommend the following changes to SSL:
\squishlist	
	\item \textbf{Smaller resolution}. Fine-grained details in high resolution images may be unnecessary, or may even confuse the contrastive loss;
	\item \textbf{Directly perform SSL on the target domain}, even when there are only \emph{a small set of training images};
	\item \textbf{Partial backbone}. As will be further analyzed, removing the last residual block in the SSL pretrained model is helpful in improving accuracies for small data. We do \emph{not} always need to train the full backbone model in SSL.
\squishend

In short, in this paper we propose a new paradigm for training SSL models, which moves away from the supervised setup: use smaller resolution, fewer training data, and only part of the pretrained model. Because all these changes are scaled-down versions of the supervised learning setup, we call it \textit{scaled-down self-supervised learning} (S3L). With S3L, obviously we can \emph{greatly accelerate self-supervised learning}, thanks to the reduction in various dimensions. As Figure~\ref{fig:scatter} shows, S3L also leads to \emph{higher accuracy} on downstream tasks \emph{using only small data from the target task}, which \emph{leads to much higher flexibility}. S3L will be empirically verified by extensive experiments in this paper.

\section{Related Works}

To avoid time-consuming and expensive data annotations and to explore better feature representations, many self-supervised methods were proposed to learn visual representations from large-scale unlabeled images or videos. Generative approaches learn to generate or otherwise model pixels in the input space (\cite{colorization:richard:ECCV16, superresolution:ledig:CVPR17, adversarial:jeff:ICLR17}). Pretext-based approaches mainly explore the context features of images or videos such as context similarity~\cite{jiasaw:mehdi:ECCV16, context:carl:ICCV15}, spatial structure~\cite{rotnet:spyros:ICLR18}, clustering property~\cite{deepclustering:caron:ECCV18}, temporal structure~\cite{sorting:lee:ICCV17}, etc.

Unlike generative and pretext-based models, contrastive learning is a discriminative approach that aims at grouping similar samples closer and diverse samples far from each other. Contrastive learning methods greatly improve the performance of representation learning, which has become the driving force of self-supervised representation learning in recent years (\cite{moco:kaiming:CVPR20, simclr:hinton:ICML20, memorybank:wu:CVPR18, byol:grill:NIPS20, swav:caron:NIPS20, simsiam:kaiming:arxiv2020}). Following MoCo~\cite{moco:kaiming:CVPR20}, contrastive learning can be viewed as a dictionary lookup task. For each encoded query $q$, there is a set of encoded keys $\{k_0,k_1,\dots\}$, among which a single positive key $k_{+}$ matches the query $q$ (generated from different views). A contrastive loss function InfoNCE~\cite{InfoNCE:arxiv2018} is employed to pull $q$ close to $k_{+}$ while pushing it away from other negative keys:
\begin{equation}
	\label{eq:contrastive}
	L_q = - \log \frac{\exp(q\cdot{k_{+}}/\tau) } {\exp(q\cdot{k_{+}}/\tau)+\sum_{k_{-}}\exp(q\cdot{k_{-}}/\tau)} \,, 
\end{equation}
where $\tau$ denotes a temperature parameter. Both SimCLR~\cite{simclr:hinton:ICML20} and MoCo are based on Equation~\eqref{eq:contrastive}. The main difference is that SimCLR samples negative pairs from the current batch while MoCo maintains a momentum memory bank. A more radical step is made by BYOL~\cite{byol:grill:NIPS20}, which discards negative sampling in contrastive learning but achieves even better results in case a momentum encoder is used. Recently, follow-up work SimSiam~\cite{simsiam:kaiming:arxiv2020} reports surprising results that simple siamese networks can learn meaningful representations even without the momentum encoder. SwAV~\cite{swav:caron:NIPS20} takes advantages of contrastive methods without computing pairwise comparisons by enforcing consistency between cluster assignments from different views.

However, they all suffer from heavy training costs because they train \textit{the entire network} on \textit{large-scale datasets} at \textit{a large resolution} for \emph{many epochs}, which is clearly inherited from the supervised learning settings. Recently, SEED~\cite{seed:fang:ICLR21} proposes to use self-supervised knowledge distillation for SSL with small models. However, it still follows this training paradigm using large resolution on large-scale datasets. In this paper, we argue that SSL should have a different learning paradigm and we aim to scale down SSL from three aspects mentioned above, i.e., resolution, model, and data.

\section{Methods}

Our fundamental assumption is that the information encoded inside the contrastive loss is much less than that encoded in the labels via the cross entropy loss. To adapt to the reduction in information, we make a paradigm shift from previous SSL methods, which mainly include 3 aspects:

\squishlist
	\item \textbf{Large (224) $\rightarrow$ small resolution (112 or even 56)}. In the supervised setting, training with larger resolutions often yields better performance for classification~\cite{bcnn:lin:ICCV15}, object detection~\cite{mask-rcnn:he:ICCV17} and semantic segmentation~\cite{deeplabv3:arxiv2017}, in spite of much higher overhead. Also, with much more fine-grained labels, object detection and semantic segmentation often need much larger resolutions (800x600 or larger) than image classification (224x224) to get good results. However, given the very weak supervision via the contrastive loss, we expect an SSL model will \emph{not learn image details}, and \emph{a low-resolution input image is intuitively a better fit}. In the next section, we show that lower resolution in fact brings higher accuracy at much less training cost, which is \emph{the opposite of the supervised learning situation}.
	
	\item \textbf{Entire $\rightarrow$ partial backbone}. We find that SSL fails to learn deep layers (e.g., conv5) well on small data, especially for large models. This phenomenon once again confirms our conjecture: the information encoded by the contrastive loss is limited and complex models suffer more from training on small data. Hence, we propose to only train shallow layers (e.g., conv1-conv4) during SSL pretraining and then train all layers during supervised fine-tuning. It greatly improves the accuracy with less cost.
	
	\item \textbf{Large $\rightarrow$ small data}. The key is to explore the power of small data with SSL. On one hand, SSL methods are often pretrained on large-scale datasets, which brings heavy training cost. On the other hand, the paradigm of pretraining followed by downstream fine-tuning will become cumbersome in some scenarios. As aforementioned, we cannot endure such a huge training cost to pretrain 10 different models on big data. Hence, how to directly perform SSL on the target small datasets is a valuable and interesting problem.
\squishend

Combining the 3 aspects above, our scaled-down self-supervised learning (S3L) framework successfully explore the power of small data with existing SSL methods, which not only greatly accelerates the training process but also achieves higher accuracy, as shown in the next section. 

\section{Experimental Results}

We used 7 small datasets for our experiments, as shown in Table~\ref{tab:dataset-overview}. First, we demonstrate the effectiveness of small resolution on small datasets as well as the large-scale ImageNet~\cite{ILSVRC2012:russakovsky:IJCV15} in Section~\ref{sec:exp-small-resolution}. Then, we investigate the effect of removing the last residual block in Section~\ref{sec:exp-small-architecture}. Finally, we explore the power of small data in Section~\ref{sec:exp-small-data}. All our experiments were conducted using PyTorch and we used Titan Xp GPUs for ImageNet experiments and Tesla K80 GPUs for small datasets. Codes will be made publicly available. 

\begin{table}
	\caption{Statistics of the 7 small datasets used in the paper.}	\label{tab:dataset-overview}
	\centering
	\renewcommand{\arraystretch}{0.9}
	\small
	\renewcommand{\multirowsetup}{\centering}
	\begin{tabular}{l|c|c|c}
			\hline
			Datasets & \# Category & \# Training & \# Testing \\
			\hline 
			CUB200~\cite{cub200} & 200 & \phantom{0}5994 & 5794  \\
			Cars~\cite{cars} & 196 & \phantom{0}8144 & 8041 \\
			Aircrafts~\cite{aircrafts} &100 &\phantom{0}6667&3333\\
			Flowers~\cite{flowers} &102&\phantom{0}2040&6149\\
			Pets~\cite{pets} &\pt37&\phantom{0}3680&3669\\
			Dogs~\cite{dogs} &120 &12000&8580\\
			DTD~\cite{dtd} &\pt47&\phantom{0}3760&1880\\
			\hline
	\end{tabular}
\end{table} 

\subsection{Small resolution is beautiful} \label{sec:exp-small-resolution}

We first investigate the effectiveness and efficiency of small resolution on small datasets in Section~\ref{sec:resolution-small-dataset}. Then, we demonstrate that small resolution is also useful for the large-scale dataset ImageNet (IN) in Section~\ref{sec:resolution-large-dataset}. 

\subsubsection{Results on CUB200 and other small datasets} \label{sec:resolution-small-dataset}

We carefully compare the influence of various input resolutions during SSL pretraining using 3 typical SSL methods, namely MoCov2~\cite{mocov2:xinlei:arxiv2020}, SimCLR~\cite{simclr:hinton:ICML20} and BYOL~\cite{byol:grill:NIPS20} under both ResNet-18 and ResNet-50~\cite{resnet:he:CVPR16}. The full learning process contains two stages: pretraining and fine-tuning. We use the pretrained weights obtained by SSL for initialization and then  fine-tune networks for classification using the cross entropy loss. Note that \emph{SSL pretraining and fine-tuning are both performed \emph{only} on the target dataset}. 

For the fine-tuning stage, we fine-tune all methods for 120 epochs using SGD with a batch size of 64, a momentum of 0.9 and a weight decay of 5e-4 for fair comparisons. For the ImageNet supervised setting, the learning rate (lr) is initialized to 0.01, which is divided by 10 every 40 epochs following \cite{NRS}. For other methods, we initialize the lr to 0.1 and use the cosine learning rate decay. We also list the results using the mixup~\cite{mixup:ICLR18} strategy, where alpha is set to 1.0. For the SSL pretraining stage, we follow the same settings in the original papers and more details are included in the appendix. Experimental results are shown in Table~\ref{tab:clean-cub200-result} (and part of the results are visualized in Figure~\ref{fig:scatter}).

\begin{table*}
	\caption{Comparisons of pretraining details, total time (GPU hours using 2 Tesla K80s) and accuracy (\%) on CUB200. All are fine-tuned for 120 epochs for fair comparisons. `N/A' means that we didn't conduct ImageNet pretraining on K80 GPUs.}
	\label{tab:clean-cub200-result}
	\centering
	\renewcommand{\arraystretch}{0.9}
	\footnotesize
	\renewcommand{\multirowsetup}{\centering}
		\begin{tabular}{l|c|c|c|c|r|c|c|c}
			\hline
			\multirow{2}{*}{Backbone}      & \multicolumn{5}{c|}{pretraining} & \multicolumn{2}{c|}{Accuracy} & \multirow{2}{*}{Total time} \\
			\cline{2-8}
			&method&resolution&\#FLOPS&epochs&time&Normal&Mixup&\\
			\hline
			\multirow{24}{*}{ResNet-18}& \multicolumn{4}{c|}{ImageNet supervised} &N/A  & 76.2 &75.0&N/A\\
			\cline{2-9}
			& \multicolumn{4}{c|}{random initialize} & 0.0& 62.0&63.4&\pt1.1\\
			\cline{2-9}
			& \multirow{10}{*}{MoCov2}& \multirow{2}{*}{224} & \multirow{2}{*}{1824.54M} & 200   & 1.6 &63.7 &65.8 &\pt2.7\\
			&&  & & 800   & 6.4 &65.0 &66.3 & \pt7.5\\
			\cline{3-9}
			&& \multirow{2}{*}{112} & \multirow{2}{*}{488.40M} & 200  & 0.9 & 64.2 & 65.4 & \pt2.0\\
			&&   & &800  & 3.6 & 66.2 & 67.4 &\pt4.7\\
			\cline{3-9}
			&& 112$\rightarrow$224  & 755.63M &800$\rightarrow$200  & 5.2& 66.4 & 68.4 & \pt6.3 \\
			\cline{3-9}
			&& \multirow{2}{*}{56} & \multirow{2}{*}{130.75M} & 200  & 0.7 & 63.2 & 64.6 & \pt1.8\\
			&&   & &800  & 2.8 & 66.1 & 67.5 & \pt3.7\\
			
			\cline{3-9}
			&& 56$\rightarrow$112  & 202.28M &800$\rightarrow$200  & 3.7 & 66.0 & 68.8 & \pt4.8 \\
			\cline{3-9}
			&& 56$\rightarrow$112$\rightarrow$224 & 295.95M &800$\rightarrow$200$\rightarrow$100  & 4.5& 66.2 & 69.3 & \pt5.6 \\
			\cline{2-9}
			& \multirow{6}{*}{SimCLR}& \multirow{2}{*}{224} &\multirow{2}{*}{1824.54M} & 200   & 1.8  &63.6&64.5&\pt2.9\\
			&&&&800&6.4&66.0&67.3&\pt8.5\\
			\cline{3-9}
			&&\multirow{2}{*}{112} &\multirow{2}{*}{488.40M} & 200   & 0.8  &64.8&67.9&\pt1.9\\
			&&&&800&3.2&67.9&69.2&\pt4.3\\
			\cline{3-9}
			&&\multirow{2}{*}{56} &\multirow{2}{*}{130.75M} & 200   & 0.6 &65.7&68.9&\pt1.7\\
			&&&&800&2.4&\textbf{68.1}&\textbf{71.0}&\pt3.5\\
			\cline{2-9}
			& \multirow{6}{*}{BYOL}& \multirow{2}{*}{224} &\multirow{2}{*}{1824.54M} & 200   &2.0   &63.2&66.0& \pt3.1 \\
			&&&&800& 8.0& 65.3&68.6 & \pt9.1\\
			\cline{3-9}
			&&\multirow{2}{*}{112} &\multirow{2}{*}{488.40M} & 200   & 0.9  &64.9&65.0&\pt2.0\\
			&&&&800&3.7&66.3&70.3&\pt4.8\\
			\cline{3-9}
			&&\multirow{2}{*}{56} &\multirow{2}{*}{130.75M} & 200   & 0.6 &64.0&67.5&\pt1.7\\
			&&&&800&2.4&67.2&70.0&\pt3.5\\
			\cline{2-9}
			
			\hline
			\multirow{25}{*}{ResNet-50}& \multicolumn{4}{c|}{ImageNet supervised} & N/A  & 81.3 &82.1&N/A\\
			\cline{2-9}
			& \multicolumn{4}{c|}{MoCov2 IN 800ep} & N/A & 77.7 & 77.9& N/A\\
			\cline{2-9}
			& \multicolumn{4}{c|}{random initialize} & 0.0 & 58.6&56.3&\pt2.1\\
			\cline{2-9}
			& \multirow{12}{*}{MoCov2}& \multirow{2}{*}{224} & \multirow{2}{*}{4135.79M} & 800   & 11.4&66.5 &62.0 &13.5\\
			&&  & & 1200   & 17.2 &69.0 &72.4 & 19.3\\
			\cline{3-9}
			&& \multirow{2}{*}{112} & \multirow{2}{*}{1091.26M} & 800  & 5.5 & 67.0 & 71.0 & \pt7.6\\
			&&   & &1200  & 8.3 & 68.9 & 74.0 & 10.4 \\
			\cline{3-9}
			&& 112$\rightarrow$224  & 1700.17M &800$\rightarrow$200  & 8.4 & 68.4 & 72.3 & 10.5 \\
			\cline{3-9}
			&& \multirow{2}{*}{56} & \multirow{2}{*}{304.06M} & 800  & 3.5 & 66.2 & 70.2 & \pt5.6\\
			&&   & &1200  & 5.3 & 68.0 & 72.3 & \pt7.4 \\
			\cline{3-9}
			&& 56$\rightarrow$112  & 461.50M &800$\rightarrow$200  & 8.4& 69.1 & 72.6 & 10.5 \\
			\cline{3-9}
			&& 56$\rightarrow$112$\rightarrow$224 & 673.14M &800$\rightarrow$200$\rightarrow$100  & 11.3& 69.8 & 72.7 & 13.4 \\
			\cline{2-9}
			& \multirow{6}{*}{SimCLR}& \multirow{2}{*}{224} &\multirow{2}{*}{4135.79M} & 200   & 3.7  &68.0 &66.5 & \pt5.8\\
			&&&&800&14.8 & 69.2& 73.0& 16.9\\
			\cline{3-9}
			&&\multirow{2}{*}{112} &\multirow{2}{*}{1091.26M} & 200   & 1.5   & 65.3& 69.8& \pt3.6\\
			&&&&800& 6.0& 71.2& \textbf{74.5}& \pt8.1\\
			\cline{3-9}
			&&\multirow{2}{*}{56} &\multirow{2}{*}{304.06M} & 200   & 0.8  &68.0 &70.9 & \pt2.9\\
			&&&&800& 3.0& \textbf{71.5}&74.2 & \pt5.1\\
			\cline{2-9}
			& \multirow{6}{*}{BYOL}& \multirow{2}{*}{224} &\multirow{2}{*}{4135.79M} & 200   & 5.2  &59.4 &57.7 & \pt7.3\\
			&&&&800&20.8 & 62.4& 64.1& 22.9\\
			\cline{3-9}
			&&\multirow{2}{*}{112} &\multirow{2}{*}{1091.26M} & 200   & 2.1  & 60.4& 62.7& \pt4.2\\
			&&&&800& 8.4& 63.3& 67.7& 10.5\\
			\cline{3-9}
			&&\multirow{2}{*}{56} &\multirow{2}{*}{304.06M} & 200   & 1.3  &63.0 &65.8& \pt3.4\\
			&&&&800& 5.0& 64.1&68.1 & \pt7.1\\
			\hline
		\end{tabular}
\end{table*}

Notice that we use the same batch size in Table~\ref{tab:clean-cub200-result} and Table~\ref{tab:cub200-longer} for all input resolutions for fair comparisons. However, we know that small resolutions require much fewer GPU memories and thus enable larger batch sizes, hence we also investigate the influence of input resolution and batch size in Table~\ref{tab:resolution-batchsize}. From these results, we have the following observations:
\squishlist
	\item \textbf{SSL pretraining is useful} and SimCLR yields the best performance among the 3 SSL methods on CUB200. MoCov2, SimCLR and BYOL all achieve much higher accuracies than random initialization when fine-tuned for 120 epochs (and similar results are obtained in Table~\ref{tab:cub200-longer} when fine-tuned for more epochs). `SimCLR 800ep (56)' (800 epochs pretraining with 56x56 input resolution using SimCLR) achieves the highest accuracy (71.0\%) for ResNet-18 and `SimCLR 800ep (112)' achieves the highest accuracy (74.5\%) for ResNet-50.
	
	\item \textbf{Small resolution achieves better performance with much fewer training cost} using various SSL methods and backbone networks. Take SimCLR 800ep as an example, 56x56 resolution achieves $5.5\%$ relative higher accuracy (71.0 v.s. 67.3) and $58.8\%$ relative fewer training hours (3.5 v.s. 8.5) than the baseline 224x224 resolution under ResNet-18.
	
	\item \textbf{Gradual transition from small to large resolution is effective} to train with large resolution for SSL methods. In Table~\ref{tab:clean-cub200-result}, we design a multi-stage pretraining strategy (gradually from small to large resolution). For example, in the `56$\rightarrow$112$\rightarrow$224' setting, the SSL pretraining process contains 3 stages: (1) train the network for 800 epochs with 56x56 input resolution; (2) then, train for 200 epochs with 112x112 resolution initialized with the weights obtained in the first stage; (3) finally, train for 100 epochs with 224x224 resolution initialized with the weights obtained in the second stage. This multi-stage training strategy achieves higher accuracy than the baseline `224' setting with fewer training hours, although the final pretraining resolution are both 224x224. It indicates that directly training with a large resolution is harmful on small datasets while a gradual transition from small to large resolution is promising to train with large resolution for SSL methods.
	
	\item As shown in Table~\ref{tab:resolution-batchsize}, \textbf{a smaller resolution enables a larger batch size}, which will further improve the performance when compared to the results in Table~\ref{tab:clean-cub200-result} where we used the same batch size for all resolutions. Take ResNet-50 as an example, the maximum batch size is 128 with 224x224 resolution using 2 K80 GPUs and a batch size of 512 will incur the `\textit{out of memory}' problem. This problem will not appear when small resolutions are used. In has already been shown that larger batch size brings higher accuracy for SimCLR in \cite{simclr:hinton:ICML20}, hence we use BYOL here for illustration in Table~\ref{tab:resolution-batchsize}. We can observe that 112x112 resolution have $12\%$ relative higher accuracy (71.8 v.s. 64.1) than the baseline 224x224 resolution, using only one-third of the time (7.0 v.s. 20.8).
\squishend

\begin{table}[t]
	\caption{Smaller resolution enables larger batch sizes (bs). All methods are trained using BYOL and ResNet-50 for 800 epochs on 2 K80 GPUs on CUB200. `N/A' indicates the 224 resolution incurs \textit{out of memory} with batch size being 512.}
	\label{tab:resolution-batchsize}
	\centering
	\renewcommand{\arraystretch}{0.9}
	\small
		\begin{tabular}{l|c|c|c|c}
			\hline
			\multirow{2}{*}{bs} & \multirow{2}{*}{resolution} & \multirow{2}{*}{time} & \multicolumn{2}{c}{Accuracy}  \\
			\cline{4-5}
			&&&Normal&Mixup\\
			\hline
			\multirow{3}{*}{128} & 224 &20.8&62.4&64.1\\ 
			&112&\pt8.3&63.3&67.7\\
			&\pt56&\pt5.0&64.1&68.1\\
			\hline
			\multirow{3}{*}{512} & 224 & N/A&N/A&N/A\\ 
			&112&\pt7.0&\textbf{67.0}&\textbf{71.8}\\
			&\pt56&\pt3.7&66.8&70.5\\
			\hline
		\end{tabular}
\end{table}

\begin{table}
	\caption{Results on CUB200 with more fine-tuning (FT) epochs. `(112 FT)' means fine-tuning with 112x112 resolution. `SimCLR 800ep (112)' means 800 epochs pretraining with 112x112 resolution under SimCLR.}
	\label{tab:cub200-longer}
	\centering
	\setlength{\tabcolsep}{2.4pt}
	\renewcommand{\arraystretch}{0.9}
	\small
	\renewcommand{\multirowsetup}{\centering}
		\begin{tabular}{l|c|c|c|c|c}
			\hline
			\multirow{2}{*}{pretraining method} & FT& \multicolumn{2}{c|}{ResNet-18} &\multicolumn{2}{c}{ResNet-50} \\
			\cline{3-6}
			&epochs&acc.&time&acc.&time\\
			\hline
			IN supervised &\pt120&75.0&\pt N/A&82.1&\pt N/A\\
			\hline
			\multirow{2}{*}{random init. (112 FT)}&\pt120&46.0&\pt0.7&44.2&\pt1.0\\
			&\pt480&54.3&\pt2.8&58.3&\pt4.0\\
			\hline
			\multirow{4}{*}{random init.}&\pt120&63.4&\pt1.1&56.3&\pt2.1\\
			&\pt480&69.2&\pt4.4&72.4&\pt8.4\\
			&1200&72.5&11.0&76.5&21.0\\
			&1600&72.1&14.7&77.2&28.8\\
			\hline
			MoCov2 800ep (224)&\pt480&72.5&10.8&76.5&19.8\\
			\hline
			MoCov2 800ep (112) & \pt480&73.6&\pt8.0&77.9&13.9\\
			\hline
			MoCov2 800ep (56)$\rightarrow$ & \multirow{2}{*}{\pt480} &\multirow{2}{*}{74.2}&\multirow{2}{*}{\pt9.9}&\multirow{2}{*}{78.4}&\multirow{2}{*}{19.7}\\
			200ep (112)$\rightarrow$100ep (224) &&&&&\\
			\hline
			SimCLR 800ep (224)	&\pt480 &73.2&10.8&77.4&23.2\\
			\hline
			SimCLR 800ep (112) &\pt480 &74.1&\pt7.6&\textbf{79.5}&14.4\\
			\hline
			SimCLR 800ep (56) &\pt480 &\textbf{75.8}&\pt6.8&79.4&11.4\\
			\hline
			BYOL 800ep (224) &\pt480 &73.1&12.4&73.0&23.2\\
			\hline
			BYOL 800ep (112) &\pt480 &73.8&\pt8.1&75.6&16.8\\
			\hline
			BYOL 800ep (56) &\pt480 &74.7&\pt6.8&76.0&13.4\\
			\hline
		\end{tabular}
\end{table}

As known in \cite{rethinkpretraining:he:ICCV19}, more training epochs is essential when training from scratch for object detection. Hence, we also investigate the effect of more fine-tuning epochs in Table~\ref{tab:cub200-longer} and we can have the following conclusions:
\squishlist
	\item \textbf{More epochs is essential when training from scratch on small datasets} for image classification. When randomly initialized, ResNet-18 achieves the highest accuracy of $75.8\%$ (1200 epochs) and the accuracy will not continue to improve with more epochs, and ResNet-50 achieves the highest accuracy of $77.2\%$ (1600 epochs).

	\item \textbf{Small resolution still has consistent improvements} compared to large resolution when fine-tuning for more epochs. When compared with the best performance of random initialization, `SimCLR 800ep (56)' achieves $3.3\%$ higher accuracy for ResNet-18 and $2.2\%$ higher accuracy for ResNet-50 with less than half of the training time.
	
	\item \textbf{Small resolution is useful for SSL but not for supervised learning}. When using the 112x112 resolution for supervised fine-tuning, the accuracy is much lower than using the 224x224 resolution (e.g., 54.3 v.s. 69.2 for ResNet-18 when fine-tuned for 480 epochs). In contrast, small resolution achieves much higher accuracy than large resolution for SSL. It indicates that SSL has limited information and hence fine-grained details in high resolution images may be unnecessary, or may even confuse the contrastive loss. In short, small resolution makes SSL easier to learn and it is not the case when we have supervised information (i.e., the labels).

	\item `SimCLR 800ep (112)' achieves higher accuracy than `MoCov2 IN 800ep' (79.5 v.s. 77.9) when fine-tuned for 480 epochs under ResNet-50. It indicates that \textbf{directly performing SSL on the target domain is promising}, even when there are only a small set of training images. Note that the latter is pretrained on ImageNet.
\squishend

\begin{figure}
	\centering
	\subfloat[ResNet-18]{ \includegraphics[width=0.8\columnwidth]{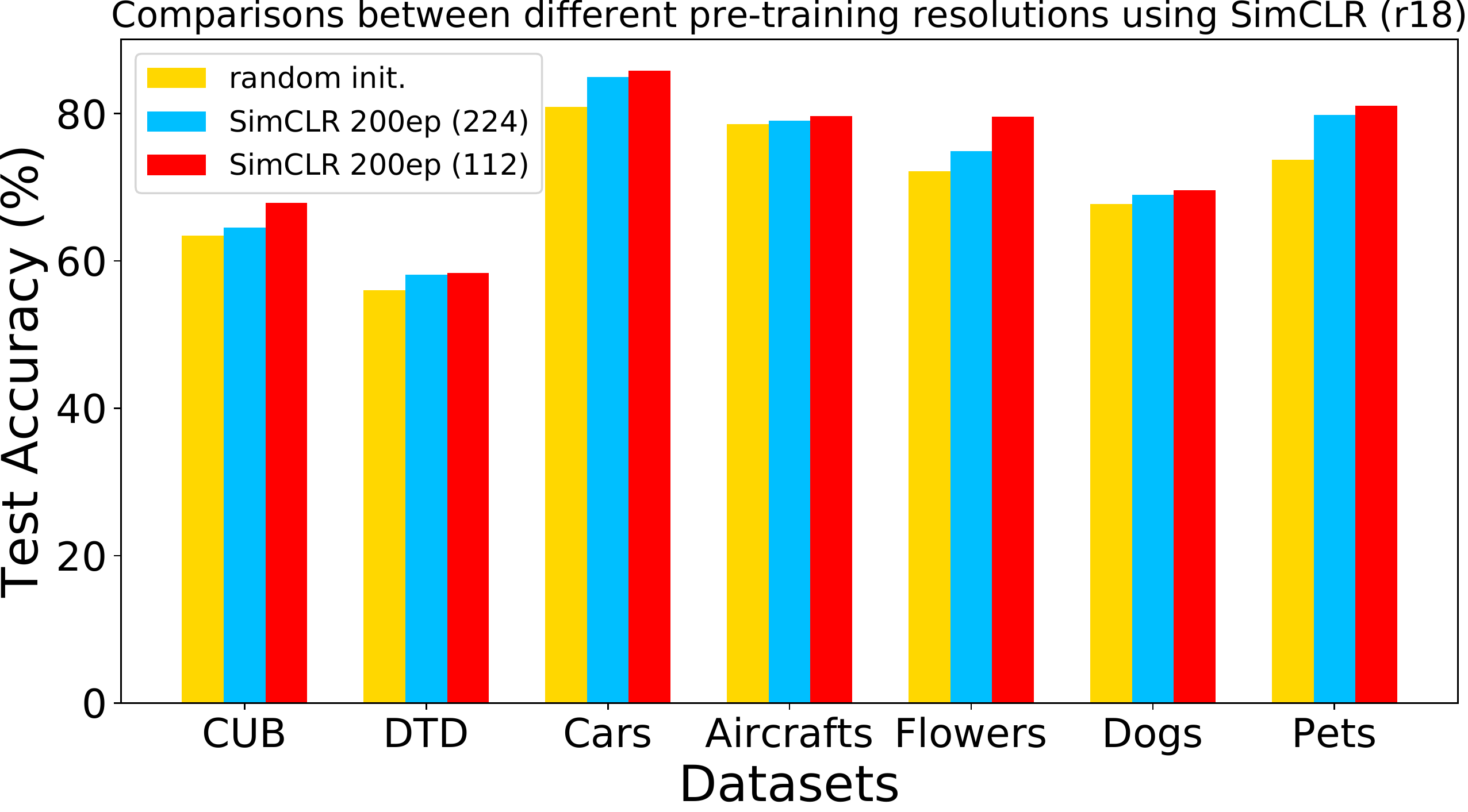} \vspace{-4pt}} \\ \vspace{4pt}
	\subfloat[ResNet-50]{ \includegraphics[width=0.8\columnwidth]{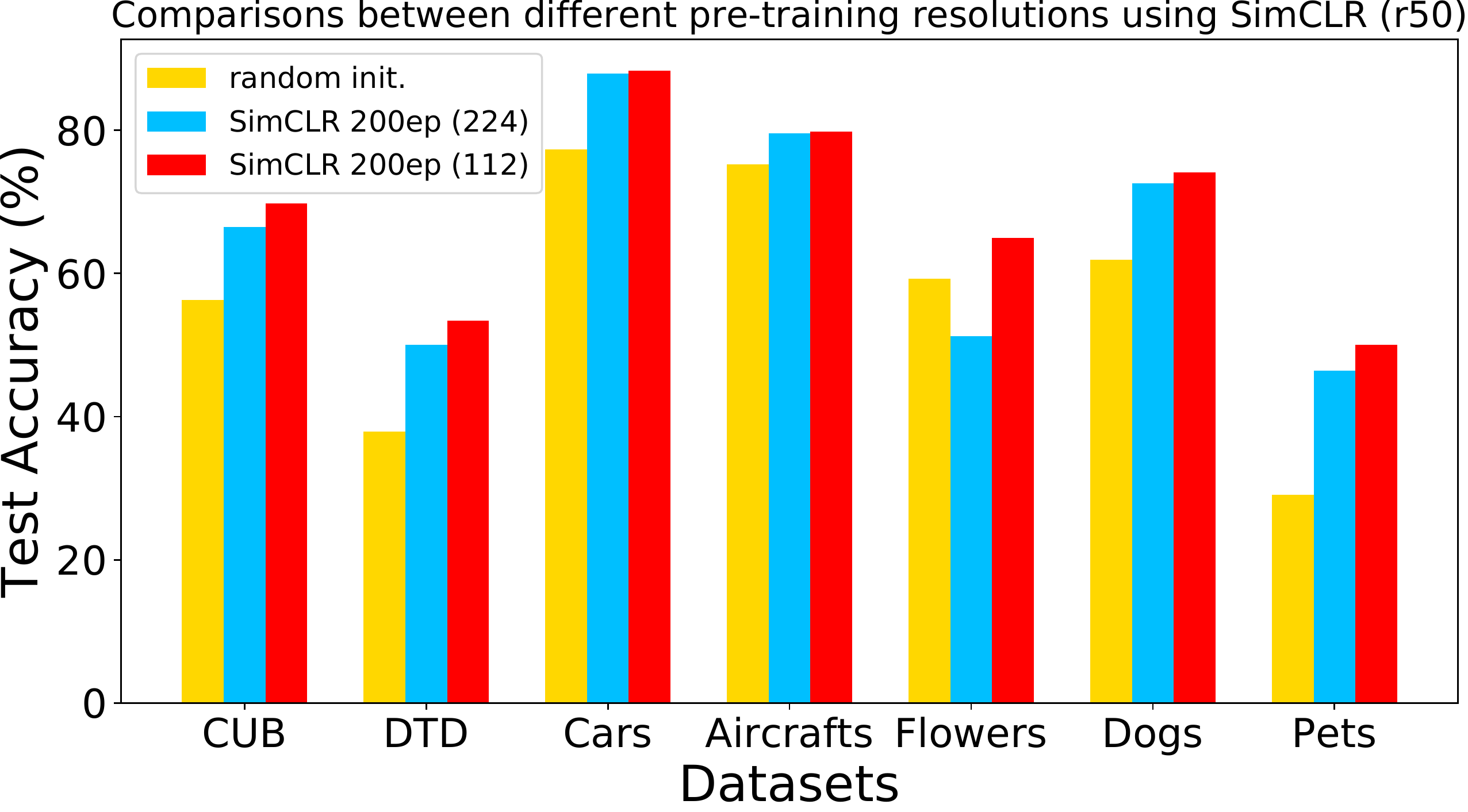} \vspace{-4pt}}
	\caption{Comparisons between different resolutions during pretraining under SimCLR on 7 small datasets.}
	\label{fig:resolution}
\end{figure}

We experimented with all the datasets in Table~\ref{tab:dataset-overview} and more results are shown in Figure~\ref{fig:resolution} (c.f. appendix for precise values). We can find consistent improvements of small resolution on all these datasets with much less training costs. 

Moreover, we visualize the feature maps using Grad-CAM~\cite{gradcam:ICCV17} to better understand the learned representations with different pretraining resolutions in the appendix. Small resolution also achieves better localization performance.

\begin{table*}
	\setlength{\tabcolsep}{4pt}
	\caption{Object detection and instance segmentation fine-tuned on COCO: bounding-box AP ($\text{AP}^{\text{bb}}$) and mask AP ($\text{AP}^{\text{mk}}$) evaluated on val2017. Pretraining time (GPU hours) was counted using 8 Titan XP GPUs.}
	\label{tab:coco-result}
	\renewcommand{\arraystretch}{0.9}
	\centering
		\begin{subtable}[b]{.625\textwidth}
			\footnotesize
			\begin{tabular}{l|r|c c c|c c c}
				\multicolumn{2}{c|}{pretraining} & \multirow{2}{*}{$\text{AP}^{\text{bb}}$} & \multirow{2}{*}{$\text{AP}_{50}^{\text{bb}}$} & \multirow{2}{*}{$\text{AP}_{75}^{\text{bb}}$} & \multirow{2}{*}{$\text{AP}^{\text{mk}}$} & \multirow{2}{*}{$\text{AP}_{50}^{\text{mk}}$} &\multirow{2}{*}{$\text{AP}_{75}^{\text{mk}}$} \\
				\cline{1-2}
				method & time &&&&&&\\
				\hline
				random init. &0&31.0&49.5&33.2&28.5&46.8&30.4\\
				IN supervised &27.8&38.4&59.2&41.6&35.0&55.9&37.1\\
				\hline
				MoCov2 200ep (224)&83.6&39.0&59.5&42.4&35.6&56.6&38.0\\
				MoCov2 200ep (112) &47.3&38.9&59.3&42.6&35.2&56.4&37.7\\
				200ep (112)$\rightarrow$ 50ep (224)&68.2&38.9&59.8&42.9&35.8&56.9&38.3\\
				\hline
				MoCov2 800ep (224) &334.4&\textbf{39.5}&59.8&43.2&\textbf{36.0}&56.9&\textbf{38.6}\\
				MoCov2 800ep (112) &189.2&39.3&59.8&42.9&35.8&56.9&38.3\\
				800ep (112)$\rightarrow$200ep (224)&272.8&\textbf{39.5}&\textbf{60.1}&\textbf{43.3}&35.9&\textbf{57.1}&\textbf{38.6}\\
			\end{tabular}
			\caption{R50-FPN (1x)}
		\end{subtable}
		\begin{subtable}[b]{.35\textwidth}
			\footnotesize
			\begin{tabular}{c c c|c c c}
				\multirow{2}{*}{$\text{AP}^{\text{bb}}$} & \multirow{2}{*}{$\text{AP}_{50}^{\text{bb}}$} & \multirow{2}{*}{$\text{AP}_{75}^{\text{bb}}$} & \multirow{2}{*}{$\text{AP}^{\text{mk}}$} & \multirow{2}{*}{$\text{AP}_{50}^{\text{mk}}$} &\multirow{2}{*}{$\text{AP}_{75}^{\text{mk}}$} \\
				&&&&&\\
				\hline
				36.7&56.7&40.0&33.7&53.8&35.9\\
				40.6&61.3&44.4&36.8&58.1&39.5\\
				\hline
				41.2&61.8&45.1&37.6&58.9&40.3\\
				41.0&61.7&45.0&37.2&58.8&39.9\\
				41.2&61.9&45.3&37.6&59.2&40.3\\
				\hline
				\textbf{41.5}&62.0&\textbf{45.7}&37.6&58.9&40.3\\
				41.4&\textbf{62.2}&45.3&37.6&59.2&40.3\\
				\textbf{41.5}&62.1&45.2&\textbf{37.7}&\textbf{59.3}&\textbf{40.4}\\
			\end{tabular}
			\caption{R50-FPN (2x)}
		\end{subtable}
\end{table*}

\begin{table*}
	\caption{Transfer learning results from ImageNet with standard ResNet-50 architecture.}
	\label{tab:classification-result}
	\centering
	\footnotesize
	\setlength{\tabcolsep}{1.6pt}
	\renewcommand{\arraystretch}{0.9}
	\renewcommand{\multirowsetup}{\centering}
	\begin{tabular}{lccccccccccc}
		\toprule
		Method &ImageNet& VOC2007 & CUB200 &Cars & Aircrafts & CIFAR10 & CIFAR100 & Caltech-101 & Flowers & Dogs&DTD\\
		\hline
		\textit{Linear evaluation:} &&&&&&&&&&&\\
		\hline
		MoCov2 200ep (224) & 67.7&80.6 &17.8&14.1&12.3&56.4&26.0&80.8&68.5&42.1&64.9\\
		MoCov2 200ep (112) & 65.3& 76.4&11.5&10.8&\pt9.5&50.1&18.4&73.4&64.6&28.3&60.1\\
		MoCov2 200ep (112)$\rightarrow$50ep (224)&66.7 & 81.2&18.8&16.2&14.7&57.8&25.5&81.8&73.5&42.9&65.1\\
		
		MoCov2 800ep (224) &\textbf{71.1}&82.8&17.5&13.4&11.8&56.8&23.6&82.1&67.8&46.4&65.2\\
		MoCov2 800ep (112) &68.4&79.0&11.0&\pt9.8&\pt8.2&53.1&21.4&72.9&56.9&36.4&61.4\\
		MoCov2 800ep (112)$\rightarrow$200ep (224)&69.8&\textbf{82.9}&18.6&14.6&12.6&\textbf{59.2}&\textbf{27.8}&82.4&72.4&46.7&65.4\\
		IN supervised&-&73.9&\textbf{61.7}&\textbf{47.1}&\textbf{23.7}&58.0&27.3&\textbf{89.1}&\textbf{86.9}&\textbf{82.2}&\textbf{68.2}\\
		\hline
		\textit{Fine-tuned:} &&&&&&&&&&&\\
		\hline
		MoCov2 200ep (224) & 73.9& 85.6&75.5&89.2&86.5&89.7&65.0&89.2&95.7&76.6&68.6\\
		MoCov2 200ep (112) & 73.6& 86.0 &76.4&89.1&86.3&90.3&67.4&89.7&94.8&76.5&68.2\\
		MoCov2 200ep (112)$\rightarrow$50ep (224)& 73.9& 86.2&77.5&88.7&86.7&89.4&65.2&91.0&95.9&77.6&70.2\\
		MoCov2 800ep (224) &75.3&87.4&77.7&89.9&87.5&89.2&65.3&89.5&95.4&77.7&67.7\\
		MoCov2 800ep (112) &75.0&87.6&77.7&88.3&86.7&\textbf{91.1}&\textbf{67.7}&90.2&95.7&77.2&68.2\\
		MoCov2 800ep (112)$\rightarrow$200ep (224)&75.3&86.2&78.8&89.1&\textbf{87.7}&90.1&66.7&91.4&96.0&77.7&69.6\\
		IN supervised&\textbf{76.1}&\textbf{89.0}&\textbf{81.3}&\textbf{90.6}&86.7&90.0&67.0&\textbf{94.1}&\textbf{96.7}&\textbf{80.1}&\textbf{74.7}\\
		\bottomrule
	\end{tabular}
	\vspace{-3pt}
\end{table*}

\begin{table}
	\caption{Object detection fine-tuned on PASCAL VOC trainval07+12. Evaluation is on test2007: $\text{AP}_{50}$ (default VOC metric), AP (COCO-style), and $\text{AP}_{75}$. All are fine-tuned for 24k iterations ($\sim$23 epochs). }
	\label{tab:voc-result}
	\centering
	\footnotesize
	\setlength{\tabcolsep}{1.8pt}
	\renewcommand{\arraystretch}{0.9}
	\renewcommand{\multirowsetup}{\centering}
	\begin{tabular}{l|r|ccc|ccc}
			\multicolumn{2}{c|}{pretraining}          &           \multicolumn{3}{c|}{R-50 FPN}           &            \multicolumn{3}{c}{R-50 C4}      \\ 
			\cline{1-8}
			method & time & $\text{AP}_{50}$ & $\text{AP}$ & $\text{AP}_{75}$ & $\text{AP}_{50}$ & $\text{AP}$ & $\text{AP}_{75}$ \\ \hline
			random init.                           &   0.0&     63.0             &     36.7        &       36.9           &   60.2               &    33.8         &         33.1         \\
			IN supervised                                &   27.8  &       80.8       &    53.5     &       58.4       &      81.3            &    53.5         &      58.8            \\ \hline
			MoCov2 200ep (224)                &     83.6     &       \textbf{81.8}       &    55.0     &       60.5       &    82.2              &      57.1       &         \textbf{64.5}         \\
			MoCov2 200ep (112)                &    47.3    &       81.3       &    54.1     &       59.5       &       82.1           &       56.8      &      63.1            \\
			200ep (112)$\rightarrow$50ep (224)   & 68.2&       81.7       &    54.8     &       60.4       &   82.2               &   56.6          &     63.5             \\ \hline
			MoCov2 800ep (224)                       &334.4   &       81.5       &    55.0     &       61.0       &     \textbf{82.6}             &      \textbf{57.7}       &         \textbf{64.5}         \\
			MoCov2 800ep (112)                       &   189.2&       81.2       &    54.3     &       61.2       &         82.4         &   57.2          &      63.9            \\
			800ep (112)$\rightarrow$200ep (224)& 272.8&       81.7       &    \textbf{55.4}     &       \textbf{61.7}       &      82.5            &     \textbf{57.7}        &64.4 \\
		\end{tabular}
\end{table}
\subsubsection{Results on ImageNet} \label{sec:resolution-large-dataset}

Now we have shown that small resolution is beautiful for small datasets, and move on to investigating the effect of small resolution for SSL on the large-scale dataset ImageNet. We use MoCov2 for illustration following the official training and evaluation protocol in \cite{mocov2:xinlei:arxiv2020}. We carefully investigate the downstream object detection performance on COCO2017~\cite{coco:LinTY:ECCV14} in Table~\ref{tab:coco-result} and Pascal VOC07\&12~\cite{VOC:mark:IJCV10} in Table~\ref{tab:voc-result}, as well as downstream classification performance on 10 datasets in Table~\ref{tab:classification-result}. The detector is Faster R-CNN~\cite{faster-rcnn:ren:NIPS15} with a backbone of R50-FPN~\cite{FPN:kaiming:CVPR17} or R50-C4~\cite{mask-rcnn:he:ICCV17} for Pascal VOC object detection and Mask R-CNN~\cite{mask-rcnn:he:ICCV17} with R50-FPN backbone for COCO, implemented in \cite{wu2019detectron2}. For ImageNet linear evaluation, we follow the same settings in \cite{mocov2:xinlei:arxiv2020}. For ImageNet fine-tuning, we train for 30 epochs with the learning rate initialized to 0.01, which is divided by 10 every 10 epochs. For other classification benchmarks, we train the network for 120 epochs with a batch size of 64 and a weight decay of 5e-4. The learning rate starts from 10.0 for linear evaluation and 0.01 for fine-tuning and is decreased every 40 epochs.

For detection, our `800ep (112)$\rightarrow$200ep (224)' strategy (c.f. Sec.~\ref{sec:resolution-small-dataset}) has comparable accuracy as the baseline `800ep (224)' setting on both Pascal VOC and COCO2017, but using 61.6 fewer training hours. Notice that the 112 resolution reduces the training time by nearly a half although it does not get as much improvements as before in small datasets when compared to the 224 resolution.

\begin{table*}
	\caption{The effect of removing conv5 on 3 small datasets. We count the extra training time of warmup epochs in total time.}
	\label{tab:cub-conv-result}
	\centering
	\renewcommand{\arraystretch}{0.9}
	\footnotesize
	\renewcommand{\multirowsetup}{\centering}
		\begin{tabular}{l|c|c|c|c|c|c|c|c|c}
			\hline
			\multirow{2}{*}{Backbone}      & \multicolumn{5}{c|}{pretraining} &  \multicolumn{3}{c|}{Accuracy} & \multirow{2}{*}{Total time} \\
			\cline{2-9}
			&resolution&setting&\#FLOPs&epochs&time&CUB200&Pets&Flowers&\\
			\hline
			\multirow{12}{*}{ResNet-18}& \multirow{6}{*}{224} & \multirow{2}{*}{baseline}& \multirow{2}{*}{1824.54M} & 200  & \pt1.6 &  65.4&74.2&76.1 & \pt2.7\\
			&&&&800&\pt6.4&66.3&76.9&82.7&\pt7.5\\
			\cline{3-10}
			& & \multirow{2}{*}{drop conv5 weights}  & \multirow{2}{*}{1824.54M}  &200  &\pt1.6  & 66.2 &77.3&79.6 & \pt2.8\\
			&&   & &800  & \pt6.4  & 68.9&77.4&83.2&  \pt7.6\\
			\cline{3-10}
			& & \multirow{2}{*}{remove conv5}  & \multirow{2}{*}{1412.51M}  &200  &\pt1.4  & 68.7&76.9&76.9  & \pt2.6\\
			&&   & &800  & \pt5.4  &  68.7&78.5&82.9&  \pt6.6\\
			\cline{2-10}

			& \multirow{6}{*}{112} & \multirow{2}{*}{baseline}& \multirow{2}{*}{\pt488.40M} & 200  & \pt0.9 &65.8 &75.2&77.2& \pt2.0\\
			&&&&800&\pt3.6&67.4&77.7&82.8&\pt4.7\\
			\cline{3-10}
			&& \multirow{2}{*}{drop conv5 weights}& \multirow{2}{*}{\pt488.40M} & 200  & \pt0.9 & 66.8 &76.6&79.6& \pt2.1\\
			&&&&800&\pt3.6&68.1&77.3&83.2&\pt4.8\\
			\cline{3-10}
			& & \multirow{2}{*}{remove conv5}  & \multirow{2}{*}{\pt353.51M}  &200  & \pt0.8 &  69.6&77.1& 77.6&\pt2.0 \\
			&&   & &800  &  \pt3.4 & \textbf{70.3} &\textbf{79.5}&\textbf{84.0}& \pt4.6 \\
			\hline

			\multirow{12}{*}{ResNet-50}& \multirow{6}{*}{224} & \multirow{2}{*}{baseline}& \multirow{2}{*}{4135.79M} & 200  & \pt2.9  &53.0  &48.0&58.2& \pt5.0 \\
			&&&&800&11.4  &62.0 &61.1 &65.6&13.5\\
			\cline{3-10}
			&& \multirow{2}{*}{drop conv5 weights}& \multirow{2}{*}{4135.79M} & 200  & \pt2.9  & 71.1  &80.2&73.9& \pt5.2 \\
			&&&&800&11.4 &71.2 &80.9&78.2& 13.7\\
			\cline{3-10}
			& & \multirow{2}{*}{remove conv5}  & \multirow{2}{*}{3332.09M}  &200 & \pt2.4 &  70.7 &78.6&76.9& \pt4.7\\
			&&   & &800  & \pt9.6 & \textbf{72.7}  &81.1&82.3&  11.9\\
			\cline{2-10}
			
			& \multirow{6}{*}{112} & \multirow{2}{*}{baseline}& \multirow{2}{*}{1091.26M} & 200  & \pt1.4  & 54.0 &48.6& 64.0 & \pt3.5 \\
			&&&&800&\pt5.5 & 71.0&61.3&67.4&\pt7.6 \\
			\cline{3-10}
			&& \multirow{2}{*}{drop conv5 weights}& \multirow{2}{*}{1091.26M} & 200  & \pt1.4  &69.9 &78.3&74.4 & \pt3.7 \\
			&&&&800&\pt5.5 & 72.2&81.1&81.6&\pt7.8 \\
			\cline{3-10}
			& & \multirow{2}{*}{remove conv5}  & \multirow{2}{*}{\pt832.06M}  &200  & \pt1.2& 71.3 &80.1&78.1 &\pt3.4 \\
			&&   & &800  &\pt4.8  &  72.2&\textbf{81.8}& \textbf{83.0}  & \pt7.1 \\
			\hline
		\end{tabular}
\end{table*}

For image classification, our method achieves lower accuracy than the baseline method for ImageNet linear evaluation, which is a popular benchmark in previous works. However, \emph{it may not be an appropriate indicator} for our method, because our `800ep (112)$\rightarrow$200ep (224)' \emph{achieves higher linear evaluation accuracies} than baseline `800ep (224)' \emph{on all the 10 downstream classification datasets}. Moreover, `800ep (112)$\rightarrow$200ep (224)' achieves higher fine-tuning accuracies than baseline `800ep (224)' on 7 out of the 10 downstream datasets and the same accuracy when fine-tuning on ImageNet. \emph{Linear evaluation may not be a good SSL evaluator}.

\subsection{Small architecture is useful} \label{sec:exp-small-architecture}

\begin{table*}
	\caption{Classification results on small datasets. `448 fine-tune' means fine-tuning with 448x448 input resolution. }
	\label{tab:other-datasets-result}
	\centering
	\footnotesize
	\setlength{\tabcolsep}{5pt}
	\renewcommand{\arraystretch}{0.9}
	\renewcommand{\multirowsetup}{\centering}
		\begin{tabular}{lccccccccc}
			\toprule
			Backbone &Method&Extra data&CUB200&Cars & Aircrafts & Flowers & Pets &DTD&Dogs \\
			\hline
			\multirow{5}{*}{ResNet-18}
			&random init.&$\times$&72.1&88.1&82.9&86.9&84.5&60.1&68.3\\
			&IN super. &\checkmark&76.2&88.3&81.2&\textbf{95.6}&90.8&68.9&76.5\\
			&S3L (ours)&$\times$&75.8&90.1&89.0&91.4&86.4&63.1&70.7\\
			&IN super. 448 fine-tune &\checkmark&\textbf{81.3}&92.0&86.9&\textbf{95.6}&\textbf{92.0}&\textbf{70.8}&\textbf{79.8}\\
			&S3L 448 fine-tune (ours) &$\times$&80.1&\textbf{92.9}&\textbf{91.0}&92.7&88.0&67.4&76.1\\
			\hline
			\multirow{6}{*}{ResNet-50}
			&random init.&$\times$&77.2&88.9&87.5&86.6&81.8&55.6&69.1\\
			&IN super. &\checkmark&81.3&90.6&86.7&96.7&91.5&74.7&80.1\\
			&MoCov2 IN 800ep&\checkmark &77.7&89.9&87.5&95.4&88.8&67.7&77.7\\
			&S3L (ours)&$\times$&79.5&91.9&90.6&91.7&88.7&63.4&73.8\\
			&IN super. 448 fine-tune &\checkmark&\textbf{84.5}&93.2&91.0&\textbf{97.0}&\textbf{93.3}&\textbf{75.5}&\textbf{83.4}\\
			&S3L 448 fine-tune (ours) &$\times$&83.8&\textbf{93.4}&\textbf{92.3}&93.4&89.3&66.9&78.1\\
			\bottomrule
		\end{tabular}
\end{table*}

\begin{table}
	\caption{Downstream object detection performance when pretrained on small ImageNet under ResNet-50 backbone. Other metrics on COCO are included in the appendix.}
	\label{tab:small-imagenet-result}
	\centering
	\footnotesize
	\setlength{\tabcolsep}{1.9pt}
	\renewcommand{\arraystretch}{0.9}
	\renewcommand{\multirowsetup}{\centering}
		\begin{tabular}{c|r|r|ccc|cc}
			\multicolumn{3}{c|}{pretraining}          &           \multicolumn{3}{c|}{VOC 07\&12}           &            \multicolumn{2}{c}{COCO 2017}      \\ 
			\cline{1-8}
			method & \#images&epochs & $\text{AP}_{50}$ & $\text{AP}$ & $\text{AP}_{75}$ & $\text{AP}^{\text{bb}}$ & $\text{AP}^{\text{mk}}$  \\ \hline
			random init.                           &0&   0&     63.0             &     36.7        &       36.9           &         31.0          &          28.5          \\
			\hline
			\multirow{5}{*}{supervised}                           &1.28M &   100  &       80.8       &    53.5     &       58.4       &       38.4         &      35.0            \\
			\cline{2-8} 
			&\multirow{2}{*}{10000} &   100  &   55.9            &    31.2    &      30.8        &     28.2           &     26.2             \\
			&&   2000  &      55.7     &     29.5      &    27.0      &             27.5   &    25.7              \\
			\cline{2-8} 
			&\multirow{2}{*}{50000} &   100  &     68.4         &   39.3      &      39.3     &    31.6            &          29.1        \\
			&&   800  &     68.2       &  38.9    &       38.1    &   29.7             &     27.7             \\
			\hline
			\multirow{5}{*}{MoCov2 (112)}                           &1.28M  &    200    &       81.3       &    54.1     &       59.5       &   38.9       &      35.2      \\
			\cline{2-8} 
			&\multirow{2}{*}{10000} &   2000  &          75.1     &     47.1   &    50.3      &        35.2        &       32.3           \\
			& &   8000  &       76.5    &    48.4     &      51.9       &        35.6        &   32.5               \\
			\cline{2-8} 
			&\multirow{2}{*}{50000} &   800  &          78.5     &     51.0   &    55.4      &          36.7      & 33.5                 \\
			& &   4000  &     79.0      &     51.4    &        56.1     &       37.3         &        34.1          \\
			\hline
		\end{tabular}
\end{table}

Furthermore, now we show that small architecture is useful and removing the last residual block in the SSL pretrained model is in fact helpful in improving accuracies. In Table~\ref{tab:cub-conv-result}, we compare 3 strategies on 3 datasets to investigate the effect of removing the last residual block during SSL pretraining:

(a) baseline: we pretrain the whole network using MoCov2 and then fine-tune for 120 epochs as before.

(b) remove conv5: During the SSL pretraining process, we remove the last residual block in the ResNet (namely conv5) and only keep conv1-conv4 for pretraining. Then, during the fine-tuning process, following the previous practices for newly added modules in \cite{bcnn:lin:ICCV15}, we freeze the conv1-conv4 blocks and warmup the conv5 block for 10 epochs with learning rate 0.1 using the supervised cross entropy loss. After that, we fine-tune all the network for 120 epochs as before. We list the total training time in Table~\ref{tab:cub-conv-result}, which includes the minor extra training time for the warmup epochs.
	
(c) drop conv5 weights: we pretrain the whole network, and then drop the conv5 weights (i.e., randomly re-initialize these weights) during the fine-tuning process. We warmpup the conv5 block for 10 epochs before fine-tuning the whole network for 120 epochs as in (b) for fair comparisons. 

From Table~\ref{tab:cub-conv-result}, we have the following observations:
\squishlist
	\item \textbf{Large models suffer more from training on small data} than small models. When comparing the baseline setting, ResNet-18 achieves much higher accuracy than ResNet-50 under the same pretraining and fine-tuning epochs. It indicates that large models are much easier to overfit on small data and they suffer from learning complex parameters with limited data. Moreover, MoCov2 even achieves lower accuracy than random initialization when pretrained for 200 epochs under ResNet-50 and the situation is alleviated when pretrained for more epochs, which further demonstrates the difficulty of learning on small data with large models.

	\item \textbf{The SSL pretrained model fails to learn deep layers well on small data}, but `drop conv5 weights' and `remove conv5' are both useful, especially for the large model ResNet-50. Take ResNet-50 800ep (224) as an example, our `remove conv5' strategy achieve $17.3\%$, $32.7\%$ and $25.5\%$ relative higher accuracies than the baseline strategy on CUB200, pets and flowers, respectively.

	\item As is well known in previous works~\cite{simclr:hinton:ICML20, mocov2:xinlei:arxiv2020}, the MLP head is essential to separate representation learning from learning specific properties of the contrastive loss. Note that the `drop conv5 weights' strategy works well and hence there is a possibility that conv5 is also fitting to certain properties of the contrastive loss rather than generic image properties. But, the `remove conv5' strategy also works well and achieves even better performance than the `drop conv5 weights' strategy with less training cost. Hence, it indicates that the underlying reason is \textbf{the lack of capability to learn complex conv5 layers well with small data for SSL models}. 
\squishend

\subsection{Small data is powerful} \label{sec:exp-small-data}

Finally, we show that small data alone is also powerful: we can achieve impressive results by directly training on small datasets without the need of pretraining on large-scale datasets (e.g., ImageNet), by combining the two strategies mentioned above.

We conducted experiments on the 7 small datasets in Table~\ref{tab:other-datasets-result}. For the ImageNet supervised setting, we follow the training protocols as before. For our S3L method, we pretrain SimCLR for 800$\sim$1600 epochs by combining both small resolution and small architecture strategies and fine-tune the whole model for 480 epochs with lr initialized to 0.1. For random initialization, we train the network for longer epochs (over 800 epochs) and report the best results.

Table~\ref{tab:other-datasets-result} shows that our S3L method has achieved impressive results on these small datasets without using any extra data. S3L even achieves higher accuracy on Cars and Aircraft than the models pretrained on ImageNet (supervised or SSL), which indicates that our method is very effective to handle small data and directly training with small data is a very promising direction. Also, we achieve the state-of-the-art results on CUB200, Cars, Aircraft and pets when training from scratch to the best of our knowledge. 
	
However, the results on DTD is still far from the ImageNet supervised baseline, especially for ResNet-50. Note that DTD is a texture dataset, and textures have an important property called the self-similarity~\cite{self-similarity:eli:ICCV07}, which means that images of the same category have highly similar internal structures and textures. This property contradicts the contrastive loss which requires an image to be similar to itself and dissimilar to others.

Another interesting thing is to investigate the performance of supervised learning and SSL when using small number of images for ImageNet (small ImageNet). We randomly sample 10000 and 50000 images to construct small ImageNet. As shown in Table~\ref{tab:small-imagenet-result}, we surprisingly find that: 
\squishlist
	\item MoCov2 (112) can learn representations well and get impressive results on downstream object detection tasks even with only 10000 images. In comparison, supervised learning fails to learn meaningful representations because they are much easier to overfit on small ImageNet. It shows that SSL is useful and essential for small data, and we can get a good starting point with SSL on small data.

	\item More epochs pretraining are beneficial for SSL on small ImageNet but not for supervised learning. When training on 10000 images, the performance on downstream tasks for supervised learning even decreases when training for more epochs, which indicates that supervised learning suffers more from the overfit problem on small data.

	\item SSL may not need as much images as supervised learning because it gets much better results than supervised baseline with only a small number of training data on ImageNet. It once again verifies our motivation: we need to scale-down from various aspects for SSL.
\squishend

\section{Conclusion}
In this paper, we proposed a shift from the existing self-supervised learning paradigm to scaled-down self-supervised learning (S3L) from 3 aspects: small resolution, small architecture and small data. Various experiments show that our method obtained a significant edge over the previous learning paradigm with much less training cost, especially on small datasets. Moreover, we achieved impressive results by directly learning on small data without any extra datasets using our S3L method, which shows that our S3L is both effective and efficient, and that learning with only small data is a promising and valuable direction.

In the future, we will explore two directions. First, we will dive into deep learning with only small data. Second, we will investigate more effective and efficient methods for self-supervised learning. 


{\small
	\bibliographystyle{ieee_fullname}
	\bibliography{main}
}

\clearpage

\appendix

\section{Training details}

\subsection{SSL settings}

The training details for MoCov2, SimCLR and BYOL on CUB200 for those experimental results presented in Table 2 in the main paper are shown in Table~\ref{tab:appendix-details}.

\subsection{Data augmentations}
For SSL pre-training, we follow the data augmentation setting in SimCLR for all the 3 SSL methods, including Gaussian blur, color distortion, random horizontal flip, random resized crop, etc. For the `224' setting, we crop 224x224 patches following previous works. For the `112' (or `56') setting, we crop 112x112 (or 56x56) patches, respectively, and other transformations remain the same.

For supervised fine-tuning, the images are resized with shorter side=256, then a 224 × 224 crop is randomly sampled from the resized image with horizontal flip and mean-std normalization.

\begin{table}
	\caption{Training details for MoCov2, SimCLR and BYOL on CUB200 for experiments presented in Table 2. $\tau$ denotes the temperature parameter and $k$ denotes the size of the memory bank in MoCov2.}
	\label{tab:appendix-details}
	\centering
	\small
	\setlength{\tabcolsep}{3.5pt}
	\begin{tabular}{l|c|c|c|c|c|c}
		\hline
		\multirow{2}{*}{Method} & \multirow{2}{*}{backbone} & \multicolumn{5}{c}{Settings}  \\
		\cline{3-7}
		&&bs&lr&lr schedule&$\tau$&k\\
		\hline
		\multirow{2}{*}{MoCov2} & ResNet-18 &128&0.03&cosine&0.2&4096\\ 
		&ResNet-50&128&0.03&cosine&0.2&4096\\
		
		\hline
		\multirow{2}{*}{SimCLR} & ResNet-18 &512&0.5&cosine&0.1&-\\ 
		&ResNet-50&128&0.125&cosine&0.1&-\\
		
		\hline
		\multirow{2}{*}{BYOL{\tiny }} & ResNet-18 &512&0.5&cosine&-&-\\ 
		&ResNet-50&128&0.125&cosine&-&-\\
		\hline
	\end{tabular}
\end{table}

\section{Localization results and visualization}

\begin{figure*}
	\centering
	\includegraphics[width=2.08\columnwidth]{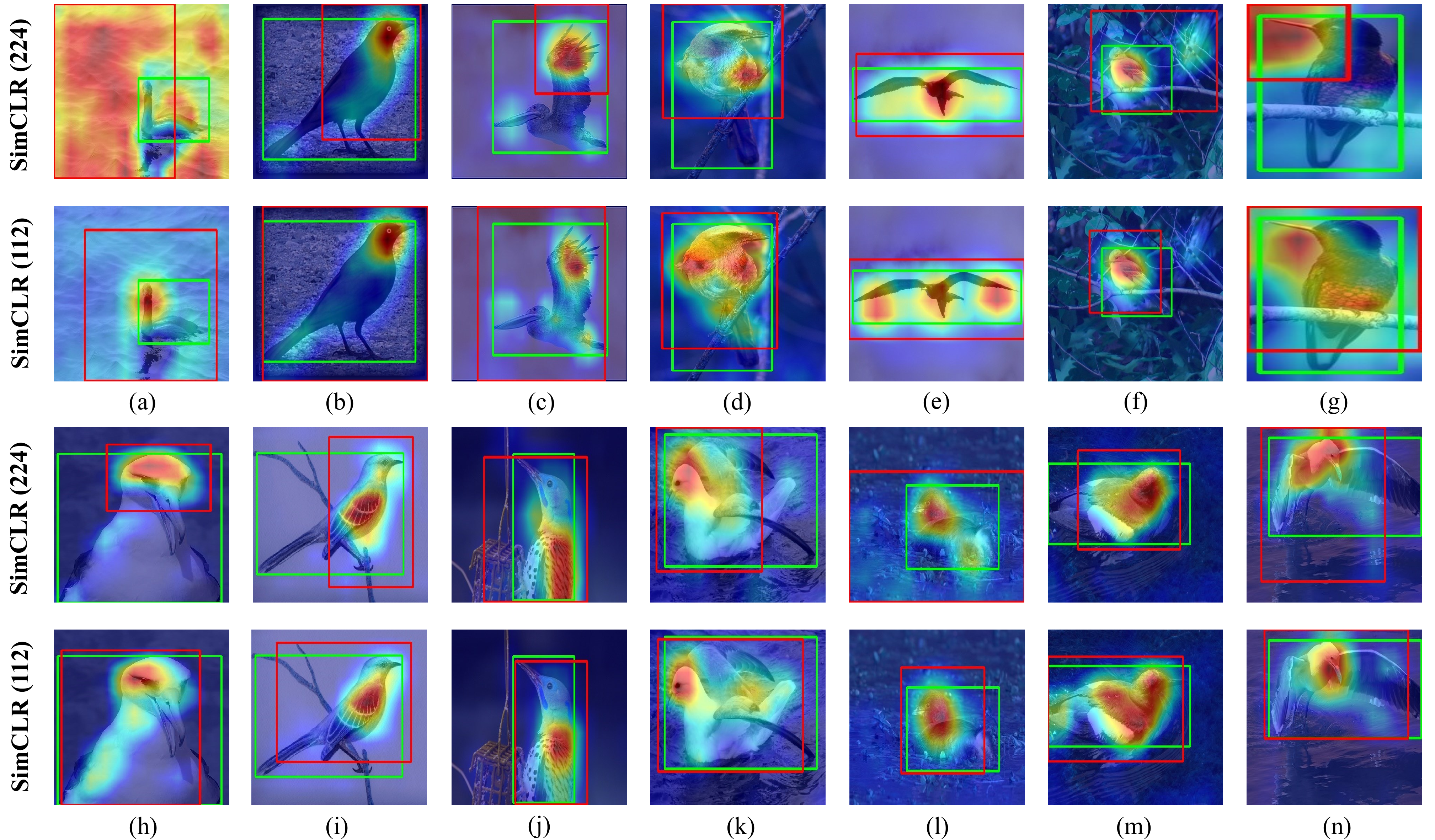}
	\caption{Comparisons of activation maps visualization between SimCLR (224) and SimCLR (112) using Grad-CAM on CUB200. Note that red boxes are Grad-CAM predicted boxes and green boxes are groundtruth boxes. This figure is best viewed in color and zoomed in.}
	\label{fig:cam}
\end{figure*}

To better understand the difference of the final learned feature representations between different input resolutions during SSL pretraining, we visualize the feature maps using Grad-CAM and evaluate the localization performance on CUB200. Following previous works, we evaluate the fine-tuned models because Grad-CAM depends on the classification head. \textit{GT-Known} Loc is correct when given the ground truth class label to the model, the intersection over union
(IoU) between the ground truth bounding box and the predicted box is 50\% or more. The localization results are shown in Table~\ref{tab:appendix-loc} and visualization results are shown in Figure~\ref{fig:cam}.

As can be seen in Table~\ref{tab:appendix-loc}, small resolution not only achieves higher classification accuracy but also achieves better localization performance. Note that we use \textit{GT known} Loc for evaluation, which is irrelevant to classification performance, hence the higher localization accuracy directly comes from better feature representations. 

\begin{table}
	\caption{Localization results using Grad-CAM on CUB200 under ResNet-50 with different SSL pretraining resolutions.}
	\label{tab:appendix-loc}
	\centering
	\small
	\setlength{\tabcolsep}{3.5pt}
	\begin{tabular}{l|c|c|c}
		\hline
		Methods & Cls acc. & \textit{GT known} Loc acc. & time \\
		\hline
		SimCLR 800ep (224) &69.2&55.9&16.9\\
		SimCLR 800ep (112) &71.2&57.5&\pt8.1\\
		SimCLR 800ep (56) &\textbf{71.5}&\textbf{57.7}&\pt5.1\\
		\hline
		
		\hline
	\end{tabular}
\end{table}

From Figure~\ref{fig:cam}, we can find that SimCLR (224) is sometimes confused by complex backgrounds (column (a) and (l)) and often localizes only the most discriminative part of an object in an image (column (c), (g), (h) and (i)). In contrast, SimCLR (112) achieves better localization results and interestingly in column (e), it successfully localizes two wings as well as the head. 

\section{More results}

The precise results of the influence of pretraining resolutions on 7 small datasets (Figure 2 in the paper) are shown in Table~\ref{tab:other-datasets-res-result}. As can be seen, small resolution achieves consistent improvements on all the 7 small datasets under both ResNet-18 and ResNet-50.

Also, we didn't list all metrics for COCO in Table 10 in Section 4.3 in the paper due to limited space, and the detailed results are shown in Table~\ref{tab:appendix-small-imagenet-result}. 

\begin{table*}[t]
	\caption{Ablation studies of different input resolutions on the 7 small datasets. All fine-tuned for 120 epochs.}
	\label{tab:other-datasets-res-result}
	\centering
	\setlength{\tabcolsep}{7pt}
	\renewcommand{\multirowsetup}{\centering}
	\begin{tabular}{lcccccccc}
		\toprule
		Backbone &Method&CUB&Cars & Aircraft & Flowers & Pets &DTD&Dogs \\
		\hline
		\multirow{3}{*}{ResNet-18}&random init. &63.4&80.9&78.6&72.2&73.7&56.0&67.7\\
		&SimCLR 200ep (224)&64.5&85.0&79.0&74.9&79.8&58.1&69.0\\
		&SimCLR 200ep (112)&\textbf{67.9}&\textbf{85.8}&\textbf{79.7}&\textbf{79.6}&\textbf{81.1}&\textbf{58.4}&\textbf{69.6}\\
		\hline
		\multirow{3}{*}{ResNet-50}&random init. &56.3&77.3&75.2&59.3&29.1&37.9&61.9\\
		&SimCLR 200ep (224)&66.5&87.9&79.6&51.2&46.4&50.0&72.6\\
		&SimCLR 200ep (112)&\textbf{69.8}&\textbf{88.3}&\textbf{79.8}&\textbf{65.0}&\textbf{50.0}&\textbf{53.4}&\textbf{74.1}\\
		\bottomrule
	\end{tabular}
\end{table*}

\begin{table*}
	\caption{Downstream object detection performance when pretrained on small ImageNet under ResNet-50 backbone. }
	\label{tab:appendix-small-imagenet-result}
	\centering
	\small
	\renewcommand{\multirowsetup}{\centering}
	\begin{tabular}{l|r|r|ccc|cccccc}
		\multicolumn{3}{c|}{pretraining}          &           \multicolumn{3}{c|}{VOC 07\&12}           &            \multicolumn{6}{c}{COCO 2017}      \\ 
		\cline{1-12}
		method & \#images&epochs & $\text{AP}_{50}$ & $\text{AP}$ & $\text{AP}_{75}$ & $\text{AP}^{\text{bb}}$& $\text{AP}^{\text{bb}}_{50}$ & $\text{AP}^{\text{bb}}_{75}$& $\text{AP}^{\text{mk}}$& $\text{AP}^{\text{mk}}_{50}$& $\text{AP}^{\text{mk}}_{75}$  \\ \hline
		random init.                           &0&   0&     63.0             &     36.7        &       36.9           &         31.0    &49.5&  33.2    &          28.5   &46.8&    30.4   \\
		\hline
		\multirow{5}{*}{supervised}                           &1.28M &   100  &       80.8       &    53.5     &       58.4       &       38.4    &59.2&  41.6   &   35.0       &55.9& 37.1    \\
		\cline{2-12} 
		&\multirow{2}{*}{10000} &   100  &   55.9            &    31.2    &      30.8        &  28.2 &46.1&29.8&26.2&43.4&27.3                  \\
		&&   2000  &      55.7     &     29.5      &    27.0      &        27.5   &45.8&28.9     &   25.7    &42.9&26.8           \\
		\cline{2-12} 
		&\multirow{2}{*}{50000} &   100  &     68.4         &   39.3      &      39.3     &   31.6     &51.0&     33.5   & 29.1 &47.9&31.0                \\
		&&   800  &     68.2       &  38.9    &       38.1    &   29.1      &49.1&  31.3     &  27.7    &45.9&    29.3        \\
		\hline
		\multirow{5}{*}{MoCov2 (112)}                           &1.28M  &    200    &       81.3       &    54.1     &       59.5       &   38.9     &59.3& 42.6 &      35.2  &56.4&37.7    \\
		\cline{2-12} 
		&\multirow{2}{*}{10000} &   2000  &          75.1     &     47.1   &    50.3      &  35.2 &55.1&38.5& 32.3& 51.9 &  34.7                \\
		& &   8000  &       76.5    &    48.4     &      51.9       &        35.6  &55.2&   38.7   &   32.5         &52.2&   34.7   \\
		\cline{2-12} 
		&\multirow{2}{*}{50000} &   800  &          78.5     &     51.0   &    55.4     &  36.7    &56.6&39.9&33.5& 53.8  & 35.8                 \\
		& &   4000  &     79.0      &     51.4    &        56.1     &       37.3     &57.7&40.7    &        34.1      &54.8& 36.6   \\
		\hline
	\end{tabular}
\end{table*}

\end{document}